# THE TRANSFERABLE BELIEF MODEL AND OTHER INTERPRETATIONS OF DEMPSTER-SHAFER'S MODEL.

Philippe SMETS[1]
IRIDIA, Université Libre de Bruxelles
50 av. F. Roosevelt, CP194/6, 1050
Brussels, Belgium.

## 1. Introduction.

Dempster-Shafer's model aims at quantifying degrees of belief. But there are so many interpretations of Dempster-Shafer's theory in the literature that it seems useful to present the various contenders in order to clarify their respective positions.

We shall successively consider the classical probability model, the upper and lower probabilities model, Dempster's model, the transferable belief model, the evidentiary value model, the provability or necessity model.

As seen none of these has received the qualification of Dempster-Shafer. In fact the transferable belief model is our interpretation not of Dempster's work but of Shafer's work as presented in his book (Shafer 1976, Smets 1988). It is a 'purified' form of Dempster-Shafer's model in which any connection with probability concept has been deleted.

[1] Research has partly been supported by the Belgian National Incentive-Program for Fundamental Research in Artificial Intelligence and the DRUMS (Defeasable reasoning and Uncertainty Management Systems) project funded by EEC grants under the ESPRIT II Basic Research Project 3085.

Any model for belief has at least two components: one **static** that describes our state of belief, the other **dynamic** that explains how to update our belief given new pieces of information. We insist on the fact that both components must be considered in order to study these models. Too many authors restrict themselves to the static component and conclude that Dempster-Shafer theory is the same as some other theory. But once the dynamic component is considered, these conclusions break down. Any comparison based only on the static component is too restricted. The dynamic component must also be considered as the originality of the models based on belief functions lies in its dynamic component.

We present a short summary of each theory.

## 2. The probability model.

In probability theory, the **static component** consists of the assessment of a probability density $p$ on the elements of $\Omega$ such that $p: \Omega \to [0, 1]$, $\sum_{\omega \in \Omega} p(\omega) = 1$.
Degrees of belief on subsets of $\Omega$ are quantified by a probability distribution $P: 2^\Omega \to [0, 1]$ such that $\forall \omega \in \Omega$, $P(\{\omega\}) = p(\omega)$ and $\forall A, B \subseteq \Omega$ with $A \cap B = \emptyset$, $P(A \cup B) = P(A) + P(B)$

The **dynamic component** is the conditioning rule: when you learn that $B \subseteq \Omega$ is true (and if $P(B) \neq 0$), $P$ is updated into the conditional probability distribution $P(.|B)$ defined on $2^\Omega$ as $P(A|B) = \dfrac{P(A \cap B)}{P(B)}$

## 3. Upper and lower probabilities models

An upper and lower probabilities model is identical to the probability model except inasmuch as it acknowledges that some probabilities might be unknown. Let $\Pi$ be the set of all those probability distributions compatible with the available information. Instead of building a meta-probability distribution on $\Pi$ as strict bayesians would recommend, one considers critical values - usually the extremes - of the various probabilities one is interested in. Various forms of partially known probability models can be described. Often $\Pi$ is a convex set of probability distributions (Kyburg 1987) uniquely defined through its upper and lower probabilities functions $P^*$ and $P_*$ where

$$\forall A \subseteq \Omega \quad \begin{array}{l} P^*(A) = \sup_{P \in \Pi} P(A) \\ P_*(A) = \inf_{P \in \Pi} P(A) \end{array}$$

When $\Pi$ contains only one element, the model reduces itself into the classical probability model.

In the classic upper and lower probabilities model, the **static component** consists in defining the upper probability distribution $P^*$ or the lower probability distribution $P_*$, both from $2^\Omega$ to $[0, 1]$. Both approaches are identical as $P^*(A) = 1 - P_*(\overline{A}) \; \forall A \subseteq \Omega$.

For each $P$ in $\Pi$, one has:
for all $A \subseteq \Omega, \; P_*(A) \leq P(A) \leq P^*(A)$

The dynamic component is the conditioning process. Conditioning on $B \subseteq \Omega$ is obtained by considering each probability distribution $P$ in $\Pi$, and conditioning them on $B$. Let $\Pi_B$ be the resulting set of conditional probability distributions:

$$\Pi_B = \{ P_B : \forall A \subseteq \Omega \; P_B(A) = P(A|B) = \frac{P(A \cap B)}{P(B)}, P \in \Pi \}$$

The upper and lower conditional probabilities functions are the upper and lower limits of these conditional probabilities:

for all $A \subseteq \Omega, P_*(A|B) = \inf_{P_B \in \Pi_B} P_B(A)$

$$= \inf_{P \in \Pi} P(A|B) = \frac{P_*(A \cap B)}{P_*(A \cap B) + P^*(\overline{A} \cap B)}$$

$$P^*(A|B) = \sup_{P_B \in \Pi_B} P_B(A)$$

$$= \sup_{P \in \Pi} P(A|B) = \frac{P^*(A \cap B)}{P^*(A \cap B) + P_*(\overline{A} \cap B)}$$

These equations have been recently studied by Planchet (1989) and Fagin and Halpern (1990).

## 4. Dempster's model.

Dempster (1967) introduced a special form of upper and lower probabilities model. For the **static component** of the model, he considers a space $X$ endowed with a probability distribution $P_X$ and a mapping $M$ from space $X$ to space $2^Y$. Let $M(x)$ denotes the image of $x$ under $M$ for $x \in X$. He defines upper and lower probabilities distribution $P^*$ and $P_*$ on $2^Y$ such that for all $A \subseteq Y$:

$P_*(A) = P_X(M_*(A))$

and $P^*(A) = P_X(M^*(A))$

where $M_*(A) = \{x : x \in X, M(x) \subseteq A, M(x) \neq \emptyset\}$

and $M^*(A) = \{x : x \in X, M(x) \cap A \neq \emptyset\}$.

The functions $P_*$ and $P^*$ are a belief and a plausibility function, respectively (see §5).

Let $P_Y$ be the (unknown) probability distribution on $2^Y$ induced by $P_X$ and the

mapping M. One way to derive $P_*(A)$ consists in writing

$$P_*(A) = \inf \sum_{x \in X} P_Y(A|x).P_X(x) \quad \forall A \subseteq Y$$

where the inf is taken over all possible values of the $P_Y(A|x)$. One has $P_Y(A|x) = 1$ whenever $M(x) \subseteq A$ and anything in $[0,1]$ otherwise. The minimum is obtained by taking $P_Y(A|x) = 0$ whenever possible. Hence:

$$P_*(A) = \sum_{x: M(x) \subseteq A} P_X(x) = P_X(M_*(A)).$$

For the **dynamic component** of the model, two types of conditioning can be considered. The first is related to upper and lower probabilities theory:

$$P_*(A|B) = \inf \frac{\sum_{x \in X} P_Y(A \cap B|x).P_X(x)}{\sum_{x \in X} P_Y(B|x).P_X(x)}$$

$$= \frac{P_*(A \cap B)}{P_*(A \cap B) + P^*(\overline{A} \cap B)}$$

where the inf is taken over all possible values of the $P_Y(A|x)$. This conditioning, hereafter called the G-conditioning, corresponds to the solution described in the upper and lower probabilities model in §3. It is not the one considered by Dempster.

For the second form of conditioning on $B \subseteq Y$, one considers that the mapping $M:X \rightarrow 2^Y$ has been transformed into mapping $M_B:X \rightarrow 2^Y$ with:

$$M_B(x) = M(x) \cap B.$$

The image of each $x \in X$ is constrained to be in B. One postulates also that the information B does not modify $P_X$, i.e. $P_X(x|B) = P_X(x)$. In that case

$$P_*(A|B) = \inf \sum_{x \in X} P_{Y|B}(A|x).P_X(x)$$

$$P^*(A|B) = \sup \sum_{x \in X} P_{Y|B}(A|x).P_X(x)$$

where $P_{Y|B}(A|x) = 1$ whenever $M(x) \cap B \subseteq A$ and anything in $[0, 1]$ otherwise. Then

$$P_*(A|B) = \frac{P_*(A \cup \overline{B}) - P_*(\overline{B})}{1 - P_*(\overline{B})}$$

$$P^*(A|B) = \frac{P^*(A \cap B)}{P^*(B)}$$

what we call the D-conditioning.

Dempster-Shafer's model often corresponds to this interpretation, i.e. Dempster's model endowed with the D-conditioning rule.

An important point in Dempster's model is that one recognizes the existence of a probability distribution on Y. The statement $P_*(A) \leq P_Y(A) \leq P^*(A)$ is meaningful as $P_Y(A)$ exists even though its exact value is unknown. It is not the case with the transferable belief model (§5) where no concept of probability distribution on Y is assumed or required.

Levi (1983) strongly criticizes the assumption $P_X(x|B) = P_X(x)$. Whenever probabilities are assumed on X and Y, one must justify why the information B leaves our probability distribution on X unchanged. One way to avert this criticism would be to avoid any reference to some underlying probability distribution. This is what we try to do in our transferable belief model interpretation of Dempster-Shafer's theory.

**5. The transferable belief model**

The transferable belief model is a model unrelated to any probability assumption.

It is postulated that evidence induces us in allocating parts of some initial finite amount of belief to subsets of the frame of discernment $\Omega$. Instead of allocating these parts of belief to the singletons of $\Omega$ as in probability theory, some parts may also be allocated to subsets. Each part represents that part of our belief that supports some subset of $\Omega$ without supporting strict subsets. Should further information be available, that part of belief m(A) allocated to $A \subseteq \Omega$ might be transferred to strict subsets of A. The **static component** of the transferable belief model corresponds to this mass allocation.

One defines the **degree of belief** bel(A) given to the set A of $\Omega$ is defined as the sum of all masses that support A,
$$bel(A) = \sum_{\emptyset \neq X \subseteq A} m(X)$$
and the **degree of plausibility** function pl(A) quantifies the total amount of belief that might support A:
$$pl(A) = bel(\Omega) - bel(\overline{A}) = \sum_{X \cap A \neq \emptyset} m(X)$$

For the **dynamic component**, suppose a mass m(A) (called a **basic belief mass**) supporting a set A of $\Omega$. You learn that subset X of $\Omega$ is impossible. The basic belief mass supporting A initially now supports $A \cap \overline{X}$. So the basic belief mass m(A) is transferred to $A \cap \overline{X}$, hence the name of the model. This corresponds to Dempster's rule of conditioning.

The transferable belief model claims that beliefs are quantified by *a single number* (bel). It is not an upper and lower probabilities model. It is a form of Dempster-Shafer's model where all relations with probability theory are cancelled. It is based essentially on Shafer's work (1976), see also Smets (1988).

The transferable belief model is not a particular case or a generalization of some probability models, nor of any meta-model based on probability distributions. Any interpretation implying the existence of an underlying probability distribution is irrelevant to our approach. We only postulate the existence of the basic belief masses assigned to subsets A of $\Omega$, each expressing the support given specifically to A and that could be transferred to strict subsets of A should new pieces of evidence become available (the conditioning process).

## 6. Other interpretations of belief functions.

### 6.1. Evidentiary Value Model.

Ekelöf (1982) initially suggested a theory of evidentiary value in judicial context (see Gardenförs et al (1983) for a survey of the topic). The model is very close to Dempster-Shafer's model and the transferable belief model.

An evidentiary argument contains three components (Gardenfors 1983):
- an evidentiary theme that is to be proved
- evidentiary facts
- evidentiary mechanisms which say that en evidentiary fact is caused by an evidentiary theme.

For these authors, the probability that the evidentiary mechanism has worked given the evidentiary facts is more important judicially than the probability of the evidentiary theme given the evidentiary facts.

In some cases, the model gives the same results as Dempster's model, but counter-examples can be built up that show the two models to be different.

## 6.2. Probability of provability or necessity.

Pearl (1988) interprets Dempster-Shafer's model as a model to quantify the probability that a proposition is provable, not that a proposition is true. Ruspini (1986) considers that bel(A) is the probability that A is necessary ($\Box A$). bel(A) could be interpreted as the sum of the probabilities $p(w; w\models \Box A)$ of those worlds w where A is necessary. Both interpretations are in fact identical. They fit in with the **static component** of Dempster-Shafer's theory. Indeed if bel(A) = $p(\Box A)$, then all inequalities characterizing the belief functions are satisfied.

But the **dynamic component** has to be justified. What is the conditioning process? Is Dempster's rule of conditioning the appropriate rule for conditioning? The problem of conditioning has apparently not yet been solved. We nevertheless believe that there is an open opportunity to show that the transferable belief model is analogous to a model where bel(A) is interpreted as $p(\Box A)$ once the conditioning process is understood.

## 7. What is Conditioning.

The static component of each model is important, but the dynamic component is even more important, though too often neglected. So we shall concentrate here more specifically on the conditioning process and Dempster's rule of conditioning through the study of the paradigm[1] of the three soldiers (S1, S2 and S3) and the three posts (P1, P2 and P3). Let an army camp with three posts, only one of them must be occupied. The officer will randomly select (with probability 1/3) one of the three soldiers. Each soldier has a habit in that
- if soldier S1 is selected, he will always go to post P1 or P2,
- if soldier S2 is selected, he will always go to post P1 or P2 or P3,
- if soldier S3 is selected, he will always go to post P1.

Before the officer selects the soldier on duty, each of them writes down on a piece of paper where he will go if selected. There are therefore six possible worlds, w1 to w6, where each world corresponds to one particular post selection (see left part of table 1). As the officer can select any one of the three guards, 18 possible worlds can be defined (referred as worlds wij if soldier Sj is selected and we were in world wi).

I want to attack the camp. I know the soldiers' preferences, how the officer selects the guard on duty, that a guard has been selected, but I do not know who. My problem is to assess my belief about which post is occupied. With the transferable belief model, bel(P1) = 1/3 (S3 was selected), bel(P2) = 0, bel(P1∨P2) = 2/3

---

[1] This is an updated version of our beehive paradigm based on Hsia suggestions.

|  | post selected by each soldier | | | occupied post according to the soldier selected | | | remaining worlds after case 1 conditioning | | | remaining worlds after case 2 conditioning | | |
|---|---|---|---|---|---|---|---|---|---|---|---|---|
| world | S1 | S2 | S3 | S1 | S2 | S3 | S1 | S2 | S3 | S1 | S2 | S3 |
| w1 | P1 | P1 | P1 | P1 | P1 | P1 | P1 | P1 | P1 | P1 | P1 | P1 |
| w2 | P1 | P2 | P1 | P1 | P2 | P1 |  |  |  | P1 |  | P1 |
| w3 | P1 | P3 | P1 | P1 | P3 | P1 | P1 | P3 | P1 | P1 | P3 | P1 |
| w4 | P2 | P1 | P1 | P2 | P1 | P1 |  |  |  |  | P1 | P1 |
| w5 | P2 | P2 | P1 | P2 | P2 | P1 |  |  |  |  |  | P1 |
| w6 | P2 | P3 | P1 | P2 | P3 | P1 |  |  |  |  | P3 | P1 |
| probabilities of being selected | 1/3 | 1/3 | 1/3 |  |  |  |  |  |  |  |  |  |

**Table 1:** the set of six worlds that represent the six possible ways posts could be selected by each soldier, and the post occupied according to which soldier has been selected by the officer.

(S1 or S3 was selected)…Three cases of conditioning can then be considered.

**Case 1:** I learn (and the soldiers know it too) that post P2 is inaccessible, so the soldiers will not select P2 if they can go elsewhere. Hence the worlds w2, w4, w5 and w6 become impossible. My beliefs about which post is occupied become: bel(P1) = 2/3 (S1 or S3 was selected), bel(P3) = 0, bel(P1∨P3) = 1.

**Case 2:** I am able to observe post P2 and realize it is empty. Hence the guard on duty had not selected P2 before being assigned the job. Hence the worlds w22, w41, w51, w52 and w61 become impossible. My beliefs are identical to those in case 1.

**Case 3:** I learn that soldier S3 was not selected. I rescale the probabilities on the space (S1, S2, S3) into P(S1) = P(S2) = 1/2. Hence bel(P1) = 0, bel(P2) = 0, bel(P1∨P2) = 1/2 (S1 was selected)…

To show the difference between case 1 and case 2, suppose I learn that soldier S2 had decided that if S1 selected post P1, S2 would not select P3. Hence world w3 is impossible.

**Case 1:** Only world w1 remains possible, hence bel(P1) = 1.
**Case 2:** Delete also worlds w31, w32 and w33. The bel(P1) = 2/3 (S1 or S3 was selected), bel(P3) = 0 and bel(P1∨P3) = 1.
**Case 3:** Unchanged.

The transferable belief model is applicable in the three cases. Case 3 is purely probabilist. In case 1, a probabilist approach will probably lead to a solution similar to the one found with the transferable belief model. The real originality of the transferable belief model lies in the way case 2 is handled. Probabilists might be tempted to defend the idea that the 1/3 probabilities present in the soldier selection process should be updated. Indeed, they could contend that the fact that P2 is not occupied somehow supports the hypothesis that soldier S2 was more probably selected than soldier S1. Hence the probabilities P' updated by the knowledge that P2 is empty should be such that P'(S2)>P'(S1). That is what Levi requires in his criticisms. The answer is that no probability is built on the wij space. So the fact that there are fewer remaining possible worlds for S1 than for S2 (4 versus 3) is irrelevant.

Case 2 conditioning corresponds to Dempster's rule of conditioning. Case 1 conditioning does not, and is usually not considered in Shafer's translators paradigms. It corresponds to the conditioning on the fact that some messages that were initially considered as possible were in fact not possible. Case 3 conditioning was given for sake of completeness.

## 8. Conclusions.

Consider the mapping M between the X and $2^Y$ as presented in paragraph 4.1. The major difference between the transferable belief and probability approaches lies of course in the way we create our beliefs on Y knowing the belief on X. The transferable belief model is based on what is available and nothing else whereas the probability analysis requires the existence of a probability distribution on Y. Bayesians assume that whenever a probability distribution $P_X$ is defined on X, then one can describe a probability distribution $P_Y$ on Y where $P_Y$ satisfies the constraints induced by $P_X$.

All alternatives to the transferable belief model explicitly or implicitly accept the Bayesian assumption: the existence of probability distributions on all relevant spaces. The real difference between the transferable belief model and all its contenders lies in this assumption. Accept it and Levi's remarks are adequate. In the transferable belief model, one never requires the existence of these probability distributions. One only recognizes that if a probability distribution can be defined on some algebra, it should induce coherence constraints on the way beliefs are allocated. But never infer that a probability distribution exists on those spaces on which we vacuously extend the belief function derived from the initial probability constraints.

Claiming the existence of a probability distribution on the space on which our beliefs are assessed is in itself already an information. Should you accept it, then the upper and lower probabilities model is appropriate.

We strongly reject the following interpretation where belief functions are used instead of upper and lower probabilities. Some authors consider that Dempster-Shafer's model (i.e. belief and plausibility functions) can be used to handle cases of ill defined probabilities, cases where there is a probability function on $\Omega$ but we only know that its values for each $A\subseteq\Omega$ is contained between two limits. They claim that all that is known is a belief function bel (or equivalently a plausibility function pl as $pl(A) = 1- bel(\overline{A})$) such that
$$\forall A\subseteq\Omega \qquad bel(A) \leq P(A) \leq pl(A)$$

This might be the case but then they should justify why the lower limits are quantified by a belief function (which can be done as in Dempster-Shafer theory). But once conditioning is involved, how do they justify the use of the D-conditioning and not of the G-conditioning. These questions have to be answered before using belief functions instead of lower probabilities functions as lower limits for the intervals and before using Dempster's rule of conditioning (and Dempster's rule of combination). Too often authors mix the two theories, carelessly introducing Dempster's rule of conditioning in an upper and lower probabilities context (see Halpern and Fagin 1990). This explains why we felt it would be useful to write this paper; hopefully we have succeeded in somehow clarifying matters.